\documentclass[12pt,a4paper]{article}
\usepackage{algpseudocode}
\usepackage{amsfonts}
\usepackage{algorithm}
\usepackage{amsmath}
\usepackage{amsthm}
\usepackage{authblk}
\usepackage{fancyhdr}
\usepackage{geometry}
\usepackage{graphicx}
\usepackage{etoolbox}
\usepackage{orcidlink}

\newcommand{\R}{\mathbb{R}}

\begin{document}

\title{Gamma Mixture Modeling for Cosine Similarity in Small Language Models}

\author[ ]{Kevin Player\footnote{\textit{kplayer@andrew.cmu.edu}} \,\orcidlink{0009-0000-7180-7985}}
\affil[ ]{Software Engineering Institute}
\affil[ ]{Carnegie Mellon University}

\maketitle

\thispagestyle{fancy}

\abstract{We study the cosine similarity of sentence transformer embeddings and observe that they are well modeled by gamma mixtures. From a fixed corpus, we measure similarities between all document embeddings and a reference query embedding. Empirically we find that these distributions are often well captured by a gamma distribution shifted and truncated to $[-1,1]$, and in many cases, by a gamma mixture. We propose a heuristic model in which a hierarchical clustering of topics naturally leads to a gamma-mixture structure in the similarity scores. Finally, we outline an expectation–maximization algorithm for fitting shifted gamma mixtures, which provides a practical tool for modeling similarity distributions.}

\section{Introduction}

\begin{figure}
\centering
\includegraphics[scale=0.4]{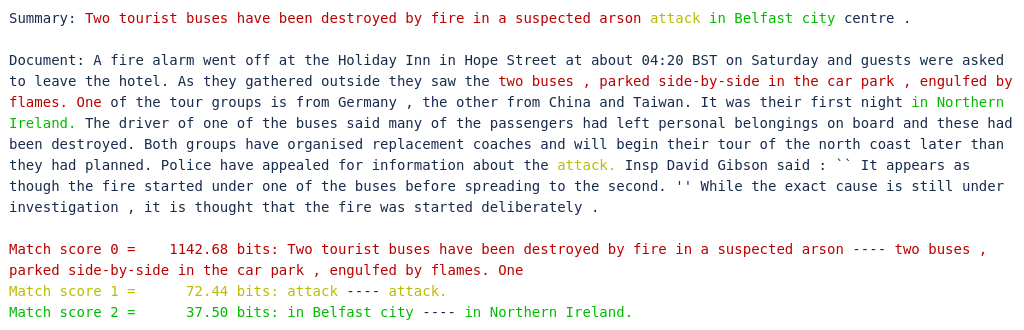}
\caption{The most significant matching of three sentence fragments in a summary(queries) with fragments in the document.  Example from xsum dataset \cite{Narayan2018DontGM} using pvalues modeled from $\texttt{all-MiniLM-L6-v2}$ \cite{Wang2020MiniLMDS}.}
\label{highlight}
\end{figure}

Cosine similarity is a widely used measure in semantic search \cite{singhal2001modern} and the advent of sentence transformers has driven widespread adoption of text-matching systems \cite{alqahtani2021survey, winastwan2024transforming}, where similarity scores are often used directly. In many semantic search applications, only the ranking of documents matters. In this paper, we instead focus on the statistical significance of a match. This perspective enables applications such as identifying the most surprising assignment of sentence fragments in a summary to fragments in a document, see an example in Figure \ref{highlight}. More generally, it can be used to compare the combined significance of multiple matches from one search result against those from another.

A common approach to computing a p-value is the permutation test \cite{moore1999bootstrapping}, which uses an empirical distribution as the null. In the context of cosine similarity, there are many examples of this approach \cite{bendidi2024benchmarking, zhou2022sense, liu2021statistically}, and specifically for sentence embeddings, see \cite{may2019measuring, wu2023transparency}. While this technique is effective, it requires a sufficiently large dataset to accurately model the tail of the null distribution and offers limited insight into its overall shape. In this paper, we propose an alternative: an accurate modeling approach that requires far less data while providing a better representation of the tail behavior.

There are several natural distributions to consider when modeling cosine similarities. Smith et al. \cite{smith2310distribution} study biological data with multivariate normals. The beta distribution (rescaled to $[-1,1]$) is another candidate, given the $[-1,1]$ support and the algebraic form of the dot product. The von Mises–Fisher (vMF) distribution is particularly appealing, as it models a Gaussian conditioned on the unit sphere $|x| = 1$, matching the normalization inherent to cosine similarity.

\begin{figure}
\centering
\includegraphics[scale=0.5]{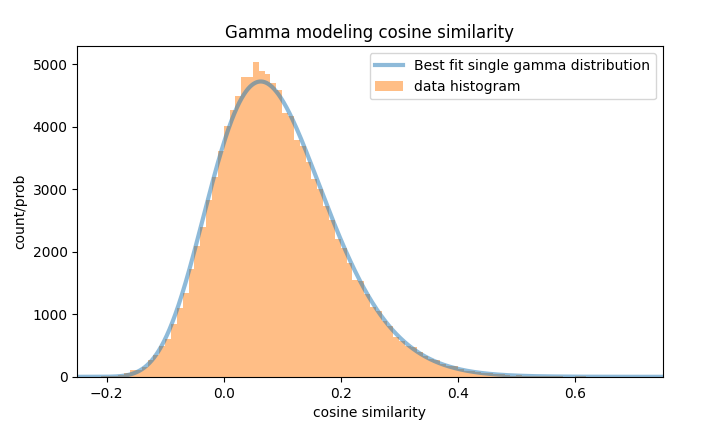}
\caption{Example distribution $D_q$ for the abstract ‘Using Genetic Algorithms for Texts Classification Problems’ (arXiv dataset). A shifted gamma distribution provides a good fit.}
\label{single_cs}
\end{figure}

However, an example histogram in Figure \ref{single_cs} shows a typical empirical distribution. It is asymmetric, with a long right tail and a nonzero mean. The normal distribution fails to capture the asymmetry, while both the beta and vMF distributions produce a heavy left tail rather than the observed right tail (see also Section \ref{sec:vmf}). Surprisingly, a simple gamma distribution, truncated and shifted to $[-1,1]$, provides an excellent fit.

Gopal and Yang \cite{gopal2014mises} introduced a hierarchical vMF mixture model for clustering. Building on this idea, we present a simplified hierarchical sampling argument that heuristically explains why cosine similarities may follow a gamma mixture distribution, see Section \ref{sec:heur}. Empirically, we find that gamma mixtures not only arise naturally from this perspective but also fit the observed data remarkably well.

In Section \ref{sec:model}, we present empirical distributions and demonstrate that they are often well modeled by gamma distributions and gamma mixtures. Section \ref{sec:math} presents a formal formulation of our gamma mixture models along with the corresponding Expectation-Maximization (EM) algorithm. Section \ref{sec:heur} presents a heuristic model based on the hierarchical clustering of topics. In Section \ref{sec:other}, we demonstrate that additional datasets and smaller language models are likewise well modeled by gamma distributions. Finally we finish in Section \ref{sec:warm} with a warm start technique for our EM and some benchmarks.

\section{Modeling Cosine Similarity} \label{sec:model}

Let $\mathcal{S}$ denote the set of all possible sentences, and let $E:\mathcal{S} \rightarrow \R^n$ be an embedding. In this paper, we mainly use the default $\texttt{BERTopic}$ sentence transformer $\texttt{all-MiniLM-L6-v2}$, $n=384$, based on MiniLM\cite{Wang2020MiniLMDS}. We focus on a {\it topical} subset $\mathcal{S}_0 \subseteq \mathcal{S}$, which consists of sentences drawn from a particular domain-specific corpus, we mainly consider the arXiv abstracts dataset \cite{CShorten2022ML-ArXiv-Papers}. Given a fixed query $q \in \mathcal{S}_0$, we study the distribution $D_q(\mathcal{S}_0)$ of
\begin{equation}
  x = {\bf cosine\_sim}(E(q), E(d)) \text{ for } d \in \mathcal{S}_0.
\end{equation}
Crucially, this restriction to $\mathcal{S}_0$ significantly affects the shape of $D_q$. In particular, the distribution tends to have a positive mean, reflecting topical coherence.

Empirically, we observe that a ($c$ - shifted) gamma distribution truncated to $[-1,1]$
\begin{equation}
  G(\alpha,c,\lambda)(x) = \frac{(x-c)^{\alpha -1} e^{-\lambda (x-c)} \lambda ^ \alpha}{\Gamma(\alpha)}
\end{equation}
often provides a good fit to the distribution of cosine similarities (see Figure \ref{single_cs}):

\begin{center}
\begin{tabular}{ccc}
  \hline\hline
  $\alpha$ & $c$ & $\lambda$ \\
  \hline
    13.3 & -0.28 & 35.5 \\
\end{tabular}
\end{center}

The need for both a shift and truncation is surprising: cosine similarity is supported on $[-1,1]$, whereas the gamma distribution is supported on $[0,\infty]$. In practice this means we are modeling with the tail-truncated portion of a shifted gamma, with most of the omitted mass lying in a highly improbable region.

\begin{figure}
\centering
\includegraphics[scale=0.5]{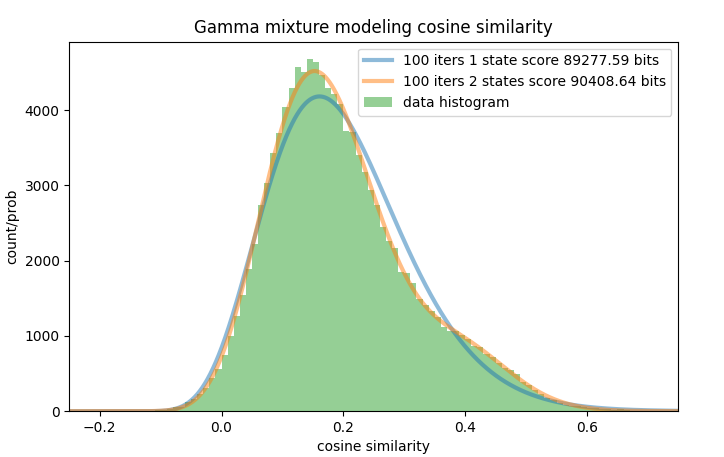}
\includegraphics[scale=0.5]{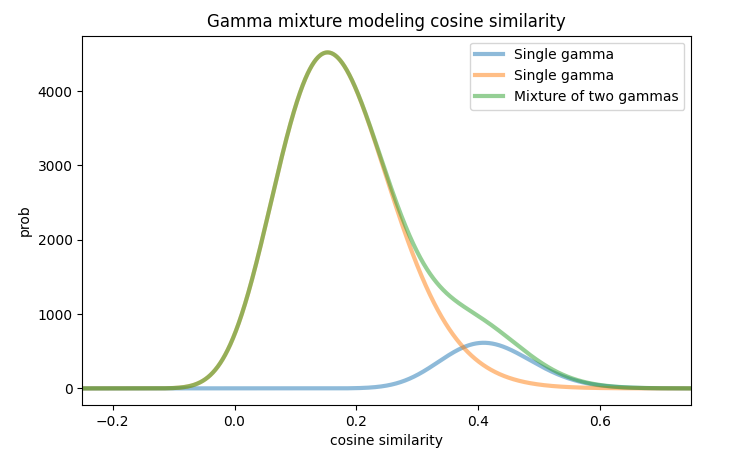}
\caption{An example of a $D_q$, $q$ in this case is the arXiv abstract for \texttt{``Why Global Performance is a Poor Metric for Verifying Convergence of Multi-agent Learning''} in the arXiv dataset. $D_q$ is fit well by a mixture of two gamma distributions.}
\label{double_cs}
\end{figure}

In cases where a single gamma distribution does not adequately capture the observed distribution, a mixture of gamma distributions often yields a better approximation. Again, the precise theoretical justification remains open, but empirical results show strong agreement with this model (see Figure \ref{double_cs}):
\begin{center}
\begin{tabular}{ccccc}
  \hline\hline
  $i$ & $\tau_i$ & $\alpha_i$ & $c_i$ & $\lambda_i$ \\
  \hline
    1 & 0.10 & 67.1 & -0.20 & 109.0 \\
    2 & 0.90 & 19.2 & -0.25 & 45.8 \\
\end{tabular}
\end{center}
where $\tau_i$ is the mixture parameter over states $i$.

\subsection{Von Mises-Fisher Modeling} \label{sec:vmf}

\begin{figure}
\centering
\includegraphics[scale=0.5]{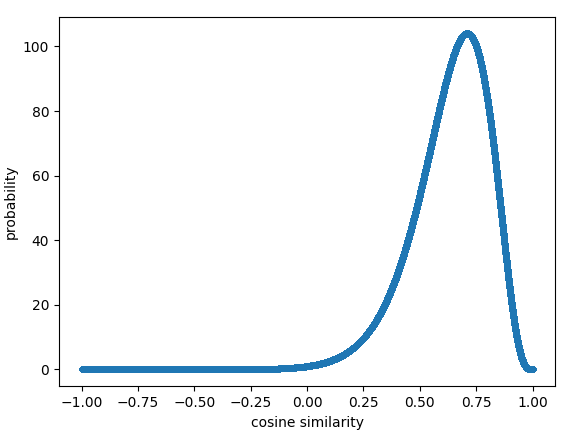}
\caption{Typical vMF distribution of cosine similarity ($d = 10$ and $\kappa = 10$).}
\label{vmffig}
\end{figure}

Gopal and Yang \cite{gopal2014mises} model semantic similarity using von Mises-Fisher vMF distribution in $d$ dimensions.  The vMF distribution has density on the sphere $|x| = 1$
\begin{equation}
  f(x) = C_d(\kappa) e ^ {\kappa \mu^T x}
\end{equation}
where $C_d(\kappa)$ does not depend on $x$, and $|\mu| = 1$.  Without loss of generality, we can pick $\mu$ to be a basis vector along the first dimension, and then cosine similarity, $t = \mu^T x \in [-1,1]$, is just the first coordinate of $x$.  We integrate along the other dimensions of $x$ to find a pdf for the cosine similarity $t$
\begin{equation}
  g(t) \propto (1 - t^2) ^ {\frac{d-3}{2}} e^{\kappa t}
\label{vmfcs}
\end{equation}
upto a constant that depends on $d$ and $\kappa$. This distribution is centered at positive $t$ but exhibits a heavy left tail, see Figure \ref{vmffig}, which conflicts with the right-tailed empirical behavior.

\section{(Shifted) Gamma Mixture Math} \label{sec:math}

\subsection{Mixture Formulation}
Suppose we are given a dataset $x_t$, and we wish to model it as a mixture of $s$ shifted\footnote{The model assumes that $x_t > c_i$} gamma distributions $G_{\alpha_i, c_i, \lambda_i}$:
\begin{equation}
P_{\alpha_i, c_i, \lambda_i}(x) = \sum_{i=1}^{s} \tau_i G_{\alpha_i, c_i, \lambda_i}(x) = \sum_{i=1}^{s} \tau_i \frac{(x-c_i)^{\alpha_i -1} e^{-{\lambda_i} (x-c_i)} {\lambda_i} ^ {\alpha_i}}{\Gamma(\alpha_i)}
\end{equation}
where $\tau_i$ are the mixture weights with $\sum \tau_i = 1$. We can use expectation maximization (EM) \cite{dempter1977maximum} to do this.

\subsection{Expectation Maximization Setup}

It is well known how to fit Gamma mixture models using Expectation Conditional\footnote{Conditional just means that we split the maximization step up into parts.} Maximization ECM \cite{young2019finite}.  We present an version for our shifted GMMs. The expectation (E) step is to compute
\begin{equation}
\gamma_{t,i} = \texttt{Prob}\left(\begin{array}{c} \text{state }i \\ \text{and} \\ \text{sample }t \end{array}\right) = \tau_i G_{\alpha_i, c_i, \lambda_i}(x_t) / P_{\alpha_i, c_i, \lambda_i}(x_t)
\end{equation}
The maximization (M) step is more involved and will fill out the remainder of this section.  We first write out the $\gamma$ weighted log likelihood function
\begin{equation}  
  Q(\alpha_i, c_i, \lambda_i) = \log(P_{\alpha_i, c_i, \lambda_i}) =\sum_{t,i} \gamma_{t,i}
\left( \begin{array}{l}
  \log \tau_i + (\alpha_i - 1)\log(x_t - c_i) \\
  - \lambda_i (x_t - c_i) + \alpha_i \log \lambda_i - \log \Gamma(\alpha_i) \\
\end{array} \right)
\end{equation}
and then take some derivatives.

\subsection{Reestimating $\tau_i$}

We compute
\begin{equation}
  \frac{\partial Q}{\partial \tau_i} = \sum_t \gamma_{t,i} \frac{1}{\tau_i}
\end{equation}
and use a Lagrange multiplier $\nu$ on $\sum \tau_i =  1$, $\frac{\partial Q}{\partial \tau_i} = \nu$, to update $\tau_i$
\begin{equation}
  \widehat{\tau_i} = \frac{\sum_t \gamma_{t,i} }{\sum_{t,j} \gamma_{t,j}}
\end{equation}

\subsection{Elimination of $\lambda_i$}

To simplify the maximization of $\alpha_i$ and $\lambda_i$, we will eliminate $\lambda_i$.  Compute
\begin{equation}
  0 = \frac{\partial Q}{\partial \lambda_i} = \sum_t \gamma_{t,i} \left(-(x_t - c) + \frac{\alpha_i}{\lambda_i}\right)
\end{equation}
and write $\lambda_i$ in terms of $\alpha_i$
\begin{equation}
  \lambda_i = \alpha_i \kappa_i
\label{plug}
\end{equation}
where
\begin{equation}
  \kappa_i = \frac{\sum_t \gamma_{t,i} }{\sum_t \gamma_{t,i} (x_t - c_i)}
\end{equation}
is the inverse of the $\gamma$-weighted mean\footnote{This mirrors the structure of the rate parameter in maximum likelihood fitting of the standard gamma distribution.} of $x_t - c_i$.
We now plug this expression back into the Q-function to obtain a reduced form:
\begin{equation}
Q_0(\alpha_i, c_i) := Q(\alpha_i, c_i, \lambda_i = \alpha_i \kappa_i),
\end{equation}
which we will maximize over $\alpha_i$ in the next step.

\subsection{Reestimating $\alpha_i$ and $\lambda_i$}

We differentiate $Q_0$ to find an equation in terms of $\alpha_i$ having eliminated $\lambda_i$
\begin{equation}
  \frac{\partial Q_0}{\partial \alpha_i} = \sum_t \gamma_{t,i} \left( \log(x_t - c_i) - \kappa_i(x_t - c_i) + \log \alpha_i + 1 + \log \kappa_i - \psi(\alpha_i)  \right)
\label{ine}
\end{equation}
where $\psi$ is the digamma function.  We can next compute
\begin{equation}
  \frac{\partial^2 Q_0}{\partial \alpha_i ^ 2} = \sum_t \gamma_{t,i} \left( \frac{1}{\alpha_i} - \psi^{(1)}(\alpha_i) \right) < 0
\end{equation}
where the positivity comes from a known trigamma inequality\footnote{This is consistent with the standard gamma distribution, where the log-likelihood is also strictly concave in $\alpha$ after eliminating $\lambda$ for the same reason.}   \cite{NIST:DLMF}. So equation (\ref{ine}) is monotone increasing and we can find a root $\widehat{\alpha_i}$ by bisecting it.  Then we use equation (\ref{plug}) to update $\lambda_i$ as $\widehat{\lambda_i} = \widehat{\alpha_i} \kappa_i$.

\subsection{Reestimating $c_i$}

Next, we focus on $c_i$
\begin{equation}
  \frac{\partial Q}{\partial c_i} = \sum_t \gamma_{t,i} \left( \frac{1-\alpha_i}{x_t - c_i} + \lambda_i \right)
\label{cbi}
\end{equation}
and the second derivative is
\begin{equation}
  \frac{\partial^2 Q}{\partial c_i^2} = \sum_t \gamma_{t,i} \frac{1-\alpha_i}{(x_t - c_i)^2}.
\end{equation}
Since $\gamma_{t,i} \ge 0$ and the denominator is always positive, the sign of the second derivative is determined entirely by $1 - \alpha_i$, which is fixed. Therefore, $Q$ is either strictly convex, linear, or strictly concave in $c_i$, depending on the sign of $1 - \alpha_i$. In particular, generically\footnote{The linear $\alpha = 1$ or $\sum_t \gamma_{t,i} = 0$ case requires no $c_i$ update.}, the equation (\ref{cbi}) has at most one solution, and we can find the root $\widehat{c_i}$ efficiently using a bisection method.

\subsection{Non-convexity of $Q_0$}

Although it might be desirable to jointly reestimate $Q_0(\alpha_i, c_i)$, the Hessian for the $t$-th summand
\begin{equation}
\begin{array}{ll}
  H &=
  \begin{bmatrix}
    \frac{\partial^2 Q}{\partial \alpha_i^2} \frac{\partial^2 Q}{\partial \alpha_ic_i} \\
    \frac{\partial^2 Q}{\partial \alpha_ic_i} \frac{\partial^2 Q}{\partial c_i^2} \\
  \end{bmatrix} \vspace{10pt}\\    
  &= \begin{bmatrix}
    \frac{1}{\alpha_i} - \psi^{(1)}(\alpha_i) & -\frac{1}{x_t - c_i} + \kappa_i \\
    -\frac{1}{x_t - c_i} + \kappa_i & \frac{1 - \alpha_i}{(x_t - c_i)^2} \\
  \end{bmatrix}
\end{array}  
\end{equation}
reveals that the objective is generally non-convex since the scaled determinant
\begin{equation}
  (x_t - c_i)^2 \det(H) = \left(\frac{1}{\alpha_i} - \psi^{(1)}(\alpha_i)\right)(1-\alpha_i) - \left(\frac{1}{x_t - c_i} + \kappa_i \right)^2
\end{equation}
is negative in the range of typical $x_t$ that we encounter.  It is an open problem to see if nonlinear fitting $Q_0$ in one step is faster than the ECM coordinate zig-zag\footnote{We currently only do one update of $\alpha_i$ and $c_i$ per EM-step, but updates could be done in any order or even repeatedly per step.}.  

\section{Hierarchical Modeling} \label{sec:heur}

We motivate the use of gamma mixture models to characterize cosine similarity. One promising perspective arises from viewing the embedding space $E$ through the lens of topic modeling \cite{grootendorst2022bertopic}, where documents and queries are associated with a hierarchy of latent topics, and embeddings are organized around their respective topic centers. Under this view, cosine similarities are naturally right-skewed, as a given query tends to be close to a cluster center in the embedding space.

\begin{algorithm}
\caption{Simulation of Hierarchical Clustering Distribution (thanks to ChatGPT for explaining my code in \LaTeX)}
\begin{algorithmic}[1]
\Require Depth $m$, Ratio $\eta$, Degree $k$, Dimension $n = 384$, Seed $s = 1$
\State Set random seed to $s$
\State Initialize $X \gets$ one vector with entries from uniform$(-1,1)$ in $\mathbb{R}^n$
\For{$i = 1$ to $m$}
    \State Initialize empty list $Y$
    \For{each vector $x$ in $X$}
        \State Sample $k$ new vectors uniformly from $(-1,1)^n$
        \State For each, compute $y \gets \eta \cdot x + \text{noise}$
        \State Append all $y$ to $Y$
    \EndFor
    \State Set $X \gets$ concatenate all vectors in $Y$
\EndFor
\For{each vector $x$ in $X$}
    \State Normalize $x$ to unit length
\EndFor
\State Let $q \gets X_0$ (the original vector)
\For{each $x$ in $X$}
    \State Compute $C \gets q \cdot x$ (cosine similarity)
\EndFor
\State \Return $C$
\end{algorithmic}
\label{alg}
\end{algorithm}

To make this intuition precise, we turn to Algorithm \ref{alg}, which constructs a hierarchical tree of cluster centers. At each iteration, a node in the binary tree splits into \texttt{degree} child nodes, centered near their parent. The centers are perturbed according to a correlation strength parameter $\eta$, creating a structured dependency across levels. This generative process induces a heavy right tail in the distribution of similarities: the cosine similarity between the query $q$ and a sampled embedding $d$ depends on their relative positions in the tree, i.e., how recently they share a common ancestor. When $d$ is drawn from a node closely related to $q$, the similarity is high; otherwise, it decays due to increasing separation in the latent space.

\begin{figure}
\centering
\includegraphics[scale=0.5]{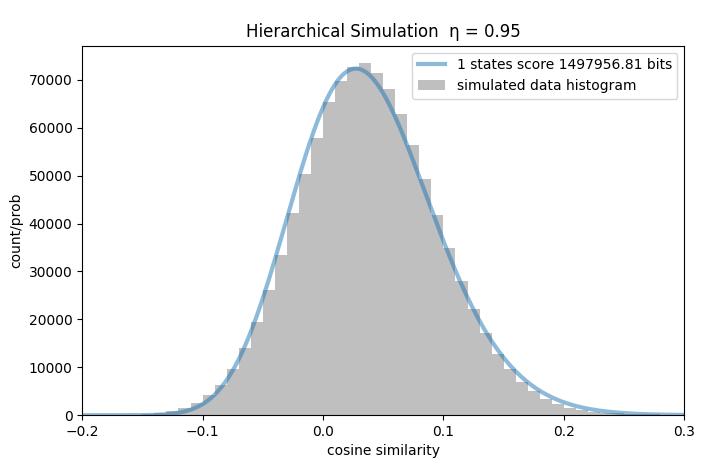}
\caption{Histogram and fitted gamma for $\eta = 0.95$ in Algorithm \ref{alg}. It is fit well by a single gamma.}
\label{hierA1}
\end{figure}

\begin{figure}
\centering
\includegraphics[scale=0.5]{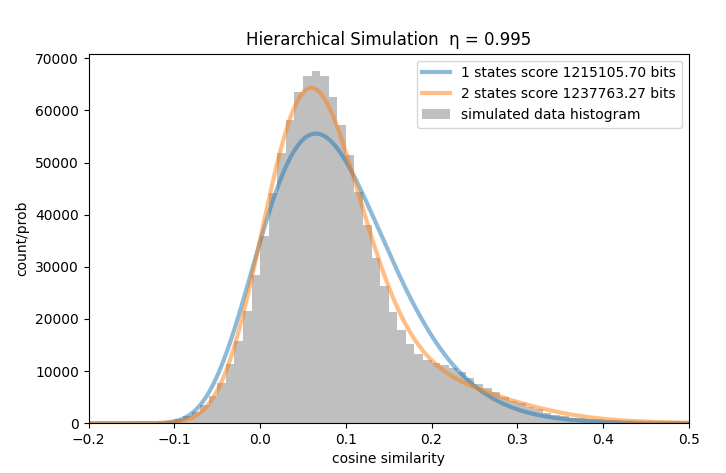}
\caption{Histogram and fitted gamma mixture for $\eta = 0.995$ in Algorithm \ref{alg}. It is fit well by a mixture of two gammas.}
\label{hierA2}
\end{figure}

With a binary tree, \texttt{degree} = 2, and sufficient \texttt{depth} (e.g., 20), we obtain enough samples to meaningfully study the distribution. For example, with $\eta = 0.95$, the distribution is well-approximated by a single gamma (see Figure \ref{hierA1}). Increasing to  $\eta = 0.995$, a mixture of two gammas fits better, see Figure \ref{hierA2}.

\subsection{Heuristic Argument -- Mixture of Multiple Hierarchies}

\begin{figure}
\centering
\includegraphics[scale=0.5]{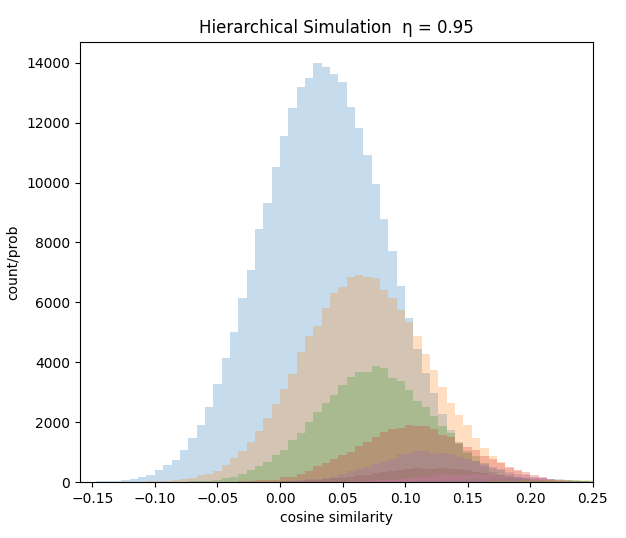}
\caption{Histograms for $\eta=0.95$ in Algorithm \ref{alg}. Colors indicate the hierarchy level whose center is closest to the query. The overlapping contributions from different levels blend to form a heavy right tail; Compare with Figure \ref{hierA1}.}
\label{hierC}
\end{figure}

We base our heuristic argument on viewing the distribution in Figure \ref{hierA1} as a mixture of level-wise contributions from the tree generated by Algorithm \ref{alg}. Each successive level contains half as many nodes and thus contributes half as many samples, producing a geometric decay clearly visible in the color-coded breakdown of Figure \ref{hierC}. Empirically, each level's component distribution is similarly shaped, shifts rightward in similarity space, and maintains comparable variance. This approximates, in the continuum, the convolution of an exponential decay with a more symmetric kernel\footnote{The kernel can be modeled as a gamma with large $\alpha$, looking approximately Gaussian in this case.}.

Averaging over latent factors -- such as depth, correlation strength, and topic context -- across multiple instances of Algorithm \ref{alg}, and their level-wise components produces a distribution shaped by overlapping exponential effects (with similar rates) and hidden variability. This blend naturally aligns with the behavior of a gamma mixture model and motivates its use as a flexible and interpretable fit.

\section{Other Models and Data} \label{sec:other}
In addition to the $\texttt{all-MiniLM-L6-v2}$ sentence embedding based on MiniLM\cite{Wang2020MiniLMDS}, we consider two other small language models based on MPNet\cite{Song2020MPNetMA} and RoBERTa\cite{Liu2019RoBERTaAR}:

\begin{center}
\begin{tabular}{rrrrr}
  \hline\hline
  model & speed & dimension & layers & context\\
  \hline
  $\texttt{all-MiniLM-L6-v2}$     & 1000  &  384  &     6  &    256 \\
  $\texttt{all-mpnet-base-v2}$    &  200  &  768  &    12  &    384 \\
  $\texttt{all-roberta-large-v1}$ &   80  & 1024  &    24  &    128 \\
\end{tabular}
\end{center}
Speed is in sentences per second on a $\texttt{Tesla V100-PCIE-16GB}$ GPU. The dimension is embedding dimension, and the context window is measured in tokens.

\begin{figure}
\centering
\includegraphics[scale=0.5]{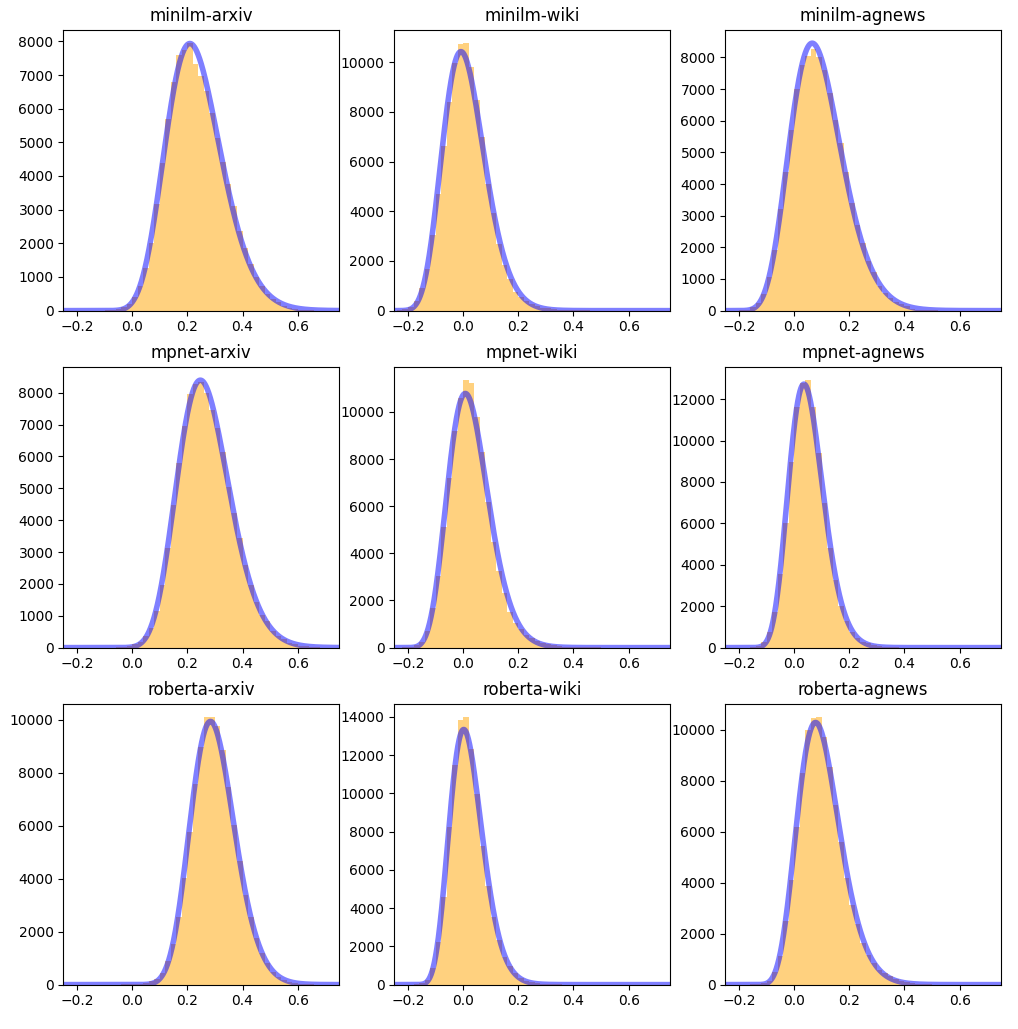}
\caption{Distribution of cosine similarity per model and dataset, each fit with a single gamma.}
\label{nine_parings}
\end{figure}

In addition to the arXiv abstracts dataset \cite{CShorten2022ML-ArXiv-Papers}, we consider Wikipedia \cite{wikidump} and ag\_news \cite{zhang2015character}.  Consider the distributions in all 9 pairings of the 3 models with the 3 datasets in Figure \ref{nine_parings}.  These are formed by taking the first sentence in each dataset and computing the cosine similarity against the first 100K other sentences. They are all pretty well described by a single gamma distribution.

\section{Warm Starting Speed up} \label{sec:warm}

We fit on a smaller set of data at first to warm start the convergence and to speed up the algorithm.  We typically do this for 95\% of the iterations on 1/20 of the data; only spending the last 5\% of the iterations on the full data set.  The difference in score is negligible, but the speed gain is an order of magnitude. Our C++ code is currently competitive with \texttt{scipy.stats.gamma.fit}:

\begin{center}
\begin{tabular}{lll}
  \hline\hline
  algorithm & states & time(ms)\\
  \hline
    scipy.stats.gamma.fit & 1 & 863\\
    ours & 1 & 116 \\
    ours & 2 & 236 \\        
    ours & 4 & 399 \\
\end{tabular}
\end{center}

\section{Acknowledgments}

{\small
\begin{verbatim}
Copyright 2025 Carnegie Mellon University.

This material is based upon work funded and supported by the
Department of Defense under Contract No. FA8702-15-D-0002 with
Carnegie Mellon University for the operation of the Software
Engineering Institute, a federally funded research and development
center.

The view, opinions, and/or findings contained in this material are
those of the author(s) and should not be construed as an official
Government position, policy, or decision, unless designated by other
documentation.

[DISTRIBUTION STATEMENT A] This material has been approved for public
release and unlimited distribution.  Please see Copyright notice for
non-US Government use and distribution.

This work is licensed under a Creative Commons
Attribution-NonCommercial 4.0 International License.  Requests for
permission for non-licensed uses should be directed to the Software
Engineering Institute at permission@sei.cmu.edu.

This work product was created in part using generative AI.

Carnegie Mellon® is registered in the U.S. Patent and Trademark Office
by Carnegie Mellon University.

DM25-1213
\end{verbatim}
}

\bibliographystyle{ieeetr}
\bibliography{bibliography}
 
\end{document}